\documentclass[a4paper]{article}

\usepackage{PRIMEarxiv}

\usepackage[utf8]{inputenc} 
\usepackage[T1]{fontenc}    
\usepackage{hyperref}       
\usepackage{url}            
\usepackage{booktabs}       
\usepackage{amsfonts}       
\usepackage{nicefrac}       
\usepackage{microtype}      
\usepackage{lipsum}
\usepackage{fancyhdr}       
\usepackage{graphicx}       
\graphicspath{{media/}}     

\usepackage{times}
\usepackage{latexsym}

\usepackage{arydshln}
\usepackage{booktabs}
\usepackage{multirow}
\usepackage{tabularx}
\usepackage{array}

\usepackage{microtype}

\usepackage{tikz}
\usepackage{cancel}
\usepackage{extarrows}
\usetikzlibrary{patterns}
\usetikzlibrary{shapes}
\usepackage{colortbl}

\usepackage[numbers]{natbib}

\title{Is Less More? Quality, Quantity and Context in Idiom Processing with Natural Language Models
}

\author{
  Agne Knietaite, Adam Allsebrook, Anton Minkov, Adam Tomaszewski,\\ \textbf{Norbert Slinko, Richard Johnson,
  Thomas Pickard, Dylan Phelps} \\
  University of Sheffield \\
  Sheffield, UK\\
  \texttt{\{tmrpickard1, drsphelps1\}@sheffield.ac.uk} \\
   \And
  Aline Villavicencio \\
  Institute for Data Science and Artificial Intelligence \\
  University of Exeter \\
  Exeter, UK\\
  \texttt{a.villavicencio@exeter.ac.uk} \\
}

\begin{document}
\maketitle

\begin{abstract}
Compositionality in language models presents a problem when processing idiomatic expressions, as their meaning often cannot be directly derived from their individual parts. Although fine-tuning and other optimization strategies can be used to improve representations of idiomatic expressions, this depends on the availability of relevant data. 
We present the Noun Compound Synonym Substitution in Books -- NCSSB -- datasets, which are created by substitution of synonyms of potentially idiomatic English noun compounds in public domain book texts. We explore the trade-off between data quantity and quality when training models for idiomaticity detection, in conjunction with contextual information obtained locally (from the surrounding sentences) or externally (through language resources).
Performance on an idiomaticity detection task indicates that dataset quality is a stronger factor for context-enriched models, but quantity also plays a role in models without context inclusion.
\end{abstract}


\section{Introduction}
\label{sec:Introduction}

Pre-trained language models such as XLNet \cite{NEURIPS2019_dc6a7e65}, GPT-3 \cite{https://doi.org/10.48550/arxiv.2005.14165} and BERT \cite{devlin-etal-2019-bert} employ principles of compositionality to encode the meaning of language, with sentence-level semantics derived from the combination of word and subword token-level elements. 
While these models perform very well on many classical text comprehension tasks \cite{raffel2023exploring, zhang-etal-2019-ernie, 10.1155/2022/1883698, hawking1988, varanasi-etal-2021-autoeqa-auto}, idiomatic language comprehension remains a challenge \cite{tayyar-madabushi-etal-2021-astitchinlanguagemodels-dataset}, as idiomatic multiword expressions (MWEs) carry non-compositional semantics \cite{constant-etal-2017-survey}. The meanings of idiomatic MWEs may therefore not be accurately captured by (sub)word composition \cite{garcia-etal-2021-probing, yu-ettinger-2020-assessing}.

Approaches to addressing this problem in the model training stages have included providing more MWE-relevant data \cite{adewumi-etal-2022-potential} and choosing specific idioms on which to train the model \cite{10.3389/frai.2022.813967}. However, it remains unclear whether quality or quantity of data is more important for idiomatic processing tasks; \citet{phelps-2022-drsphelps} provides evidence for the former, and \citet{adewumi-etal-2022-potential} the latter.

To more fully capture contextual clues that can help derive the idiomatic meaning, additional sentences from the local paragraph context can be taken into account when modeling idiomatic expressions \cite{tayyar-madabushi-etal-2021-astitchinlanguagemodels-dataset}. More global contextual information or ``commonsense" knowledge about an idiomatic expression has also been incorporated, using glosses from external knowledge sources \cite{hauer-etal-2020-ualberta} or the whole target sentence \cite{chakrabarty-etal-2022-rocket}.

As the impacts of data quality and quantity, the inclusion of external knowledge and their interactions on idiomaticity detection are unclear, we aim to investigate these factors; in particular:
\begin{itemize}
    \item Does data quality or quantity play a bigger role in model performance when identifying idiomatic language?
    \item Does additional contextual information  play a role in modeling idiomatic language?
\end{itemize}

To facilitate this, we create the Noun Compound Synonym Substitution in Books (NCSSB) datasets, which vary in size and quality and contain instances of potentially idiomatic expressions together with surrounding context sentences ($\S$\ref{sec:Datasets}).

Language models using several different context inclusion techniques are trained on these datasets ($\S$\ref{sec:Models}). The performance of these models is evaluated ($\S$\ref{sec:Results}) on the idiom representation task of SemEval 2022 Task 2 Subtask B \cite{tayyar-madabushi-etal-2022-semeval}. We conclude with discussion of our findings ($\S$\ref{sec:Discussion}). We also make our datasets\footnote{\url{https://doi.org/10.15131/shef.data.25259722}} and the code used for dataset creation and model enhancement\footnote{\url{https://github.com/agneknie/com4520DarwinProject}} publicly available.


\section{Related Work}
\label{sec:Background}

\begin{figure*}[h]
  \centering
  \begin{tikzpicture}

    \definecolor{myyellow}{RGB}{255, 229, 153}

    \node[font=\bfseries, text width=4.0cm, align=center] (textFirst) at (0,1.5) {Sentence (E)};
    \node[font=\itshape, text width=3cm, align=center] at (0,0.0) {I initially feared that taking it would make me a};
    \node[font=\itshape\bfseries, fill=myyellow, text width=2cm, align=center] at (0,-1) {guinea pig};
    
    \draw[<-, myyellow, line width=1.5pt] (2,0) -- (4,0);
    \draw[myyellow, fill=myyellow] (4,0) circle (2pt);

    \node[font=\bfseries, text width=4.0cm, align=center] at (6,1.5) {Correct\\ Replacement (E$_c$)};
    \node[font=\itshape, text width=3cm, align=center] at (6,0.0) {I initially feared that taking it would make me a};
    \node[font=\itshape\bfseries, fill=myyellow, text width=2cm, align=center] at (6,-1) {test subject};
    
    \draw[->, myyellow, line width=1.5pt] (8,0) -- (10,0);
    \draw[myyellow, fill=myyellow] (8,0) circle (2pt);

    \node[font=\bfseries, text width=4.0cm, align=center] at (12,1.5) {Incorrect\\ Replacement (E$_i$)};
    \node[font=\itshape, text width=3cm, align=center] at (12,0.0) {I initially feared that taking it would make me a};
    \node[font=\itshape\bfseries, fill=myyellow, text width=1.5cm, align=center] (textLast) at (12,-1) {animal};
    
  \end{tikzpicture}
  
  \caption{\centering SemEval Task 2 Subtask B sentence structures used in \emph{Bronze} dataset generation}
  \label{fig:bronze-generation}
\end{figure*}
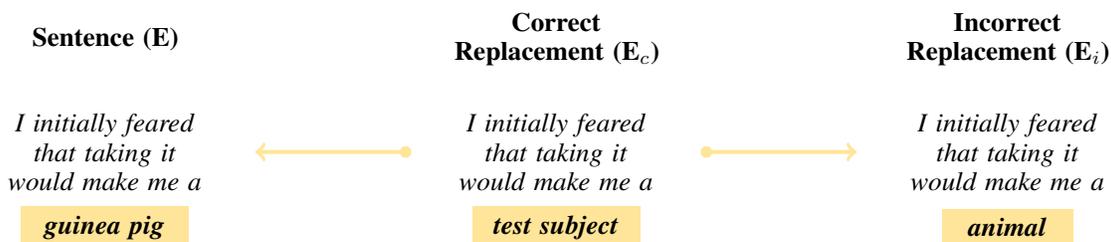

\subsection{Idiom Representation}

The capability of large language models in processing idiomatic expressions has been the focus of shared tasks like SemEval 2022 Task 2 \cite{tayyar-madabushi-etal-2022-semeval}, which focuses on idiomatic MWE representation. Subtask B requires a model to return a Semantic Text Similarity (STS) score indicating the similarity between two  sentences. The dataset comprises sentences containing a potentially idiomatic expression, denoted \emph{E}, and two variations of \emph{E}: \emph{E$_c$} is a sentence with a valid MWE replacement and \emph{E$_i$} is a sentence with an incorrect replacement given the context. See \hyperref[fig:bronze-generation]{Figure \ref*{fig:bronze-generation}}, where the idiomatic \emph{“guinea pig”} is replaced by the semantically-equivalent \emph{“test subject”} in \emph{E$_c$} and by \emph{“animal”} (which would be valid for a literal \textit{guinea pig}) in  \emph{E$_i$}.

The semantic similarity between \emph{E} and \emph{E$_c$} should be equal to 1, while the similarity between  \emph{E} and \emph{E$_i$} should be equal to the similarity between \emph{E$_c$} and \emph{E$_i$} \citep{tayyar-madabushi-etal-2022-semeval}:
\[sim(E, E_c) = 1\]
\[sim(E, E_i) = sim(E_c, E_i)\]

To perform well on both of these objectives, a model needs to produce similar embeddings of \emph{E} and \emph{E$_c$}, which would be an indication that the model was able to capture the correct meaning of the idiomatic expression.

\subsection{Idiomaticity Detection}

Subtask A of SemEval 2022 Task 2 \cite{tayyar-madabushi-etal-2022-semeval} focuses on the related task of idiomaticity detection -- binary classification of sentences according to whether or not a given expression is used in its idiomatic sense. \citet{hauer-etal-2022-ualberta} explore the idea of providing definitions of the individual component words of an expression for this task. Two glosses, sourced from BabelNet \cite{navigli-ponzetto-2010-babelnet} and the Open Multilingual Wordnet \cite{bond-foster-2013-linking}, are supplied for each individual word of the MWE in a target sentence. This method achieves an increase in performance over the baseline model of 4.1\% and 5.3\% in the zero-shot and one-shot settings respectively \cite{tayyar-madabushi-etal-2022-semeval}.

In this context, GlossBERT \cite{huang-etal-2019-glossbert} would offer a solution to the word sense disambiguation problem \cite{ide-veronis-1998-introduction}, in this case for the literal and idiomatic senses. Word senses are represented as `context-gloss pairs', each containing a sentence including a marked target word and a single definition of this word, obtained from WordNet 3.0 \cite{wordnet_1998}. Although glosses for the complete MWE would be useful, these are much less available than glosses for individual words. In an analysis of the coverage of the SemEval 2022 Task 2 dataset \cite{tayyar-madabushi-etal-2022-semeval},  81.6\% of the 223 MWEs in the training data have at least one gloss in WordNet, and of the remaining MWEs, 68.2\% are idiomatic. Wiktionary covers more of the training data, with 86.1\% of MWEs having a gloss and only 22.6\% of the remaining MWEs being idiomatic. 

Assuming the `idiom principle' \cite{sinclair:1991}, which  suggests that a person has a lexicon of pre-constructed phrases \cite{https://doi.org/10.1515/text.1.2000.20.1.29}, having access to MWE glosses would be beneficial for a model to represent potentially idiomatic expressions more accurately \cite{hashempour:2020}. Moreover, as native speakers seem to learn the meaning of idioms from being exposed to repeated use in context \cite{nunberg_1994}, a  model could also be expected to learn from being exposed to sentences containing idiomatic expressions.

For NLP tasks a huge amount of data can be crawled from the web \cite{https://doi.org/10.48550/arxiv.2005.14165}.
However, data collected from the web can be of low quality and may often include text that is not natural language \cite{raffel2023exploring}. Ignoring the quality of the data and using it to train models often leads to a drop in model performance \cite{https://doi.org/10.48550/arxiv.1806.02847}. 
One method to improve such data quality is to filter down a large dataset into a smaller but higher-quality subset; \citet{raffel2023exploring} found that the filtered Colossal Clean Crawled Corpus (C4) dataset performed better than the unfiltered variant in a range of tasks, despite being nearly 10 times smaller than the base Common Crawl.

For supervised learning tasks, any unlabelled training data gathered must be annotated, requiring costly manual annotation methods \cite{reddy-etal-2011-empirical, biemann-giesbrecht-2011-distributional, cordeiro-etal-2019-unsupervised}. For potentially idiomatic natural language data, sentences are often labelled either by language experts \cite{venkatapathy-joshi-2005-measuring, farahmand-etal-2015-multiword, savary-etal-2017-parseme} or through crowd-sourcing \cite{reddy-etal-2011-empirical, biemann-giesbrecht-2011-distributional, cordeiro-etal-2019-unsupervised}. To minimize human error and bias in the annotations, each data point is often annotated by more than one person. Various metrics are used to calculate the quality and annotator agreement of the labels, such as Spearman correlation coefficient \cite{reddy-etal-2011-empirical, farahmand-etal-2015-multiword, cordeiro-etal-2019-unsupervised}, Cohen’s kappa coefficient \cite{savary-etal-2017-parseme, tayyar-madabushi-etal-2021-astitchinlanguagemodels-dataset} and Kendall’s tau \cite{venkatapathy-joshi-2005-measuring}.

Data augmentation techniques can be used to increase dataset size by creating new synthetic data based on an initial set \cite{tidke_2022}. 
If high-quality data can be enlarged by automatic methods, one can potentially reduce reliance on costly manual methods.
One SemEval 22 Task 2 entry \cite{chu-etal-2022-hit} used the Easy Data Augmentation (EDA) approach of \citet{wei-zou-2019-eda,karimi-etal-2021-aeda-easier}, reducing data requirements by around half with no impact on performance. However, this is not always a suitable solution for MWEs, as component words can be removed as a result of EDA's random insertions and edits. Thus, we hypothesise that a more targeted form of augmentation could increase performance further \cite{10.21203}.

\section{Datasets}
\label{sec:Datasets}

    \subsection{Data Collection}
    \label{sec:DataCollection}
    
Data was collected from the Project Gutenberg corpus \cite{gutenberg}, which consists of over forty thousand English books in the public domain. 
The resulting data was tokenised into individual sentences using regular expressions. The sentences obtained were then filtered for idiomatic phrases as well as their literal and figurative synonyms found in the extended noun compound senses dataset provided by \citet{tayyar-madabushi-etal-2021-astitchinlanguagemodels-dataset}. The resulting dataset is split into three main tiers of data, with some subsets (see \hyperref[tab:dataset-summary]{Table \ref*{tab:dataset-summary}}).

        \subsubsection{Bronze Dataset}
    
Sentences in this dataset are generated by an approach outlined by \citet{tayyar-madabushi-etal-2021-astitchinlanguagemodels-dataset} \hyperref[fig:bronze-generation]{(Figure \ref*{fig:bronze-generation})}:
  
  \begin{enumerate}
     \item Sentences containing \textbf{synonyms} of the figurative meanings of idioms are found (\emph{E$_c$});
     \item Figurative meaning synonyms are replaced with idioms, forming \emph{E};
     \item Idioms are replaced by literal meanings, forming \emph{E$_i$}.
     
  \end{enumerate}

As this approach is purely automatic, it yields many entries (approx. 1,400,000) but is expected to be lacking in quality, as indicated by previous research \cite{https://doi.org/10.48550/arxiv.1806.02847, radford2019language}.
  
        \subsubsection{Silver Dataset}
  
A semi-automatic approach, with some concern for quality, where the \emph{Bronze} dataset is filtered to exclude some of the low-quality sentences \cite{raffel2023exploring}. Filtering is conducted by using constrained patterns to identify sentences where synonyms can be replaced by idioms without negatively affecting the meaning of the sentence itself. The SemEval data \cite{tayyar-madabushi-etal-2022-semeval} is taken as a high-quality dataset, whose sentences are then compared to the \emph{Bronze} dataset. The \emph{Silver 1}, \emph{Silver 5}, and \emph{Silver 10} are respectively the top 1\%, 5\% and 10\% of the \emph{Bronze} dataset when ranked according to cosine similarity with frequency count vectors of sentences for a given MWE in the SemEval dataset. The sizes of these datasets are approximately 10,000, 50,000 and 100,000 respectively\footnote{Some additional filtering of grammatically and structurally incorrect sentences is also applied.}. 

        \subsubsection{Gold Dataset}
  
A hand-labelled dataset, reviewed manually by (near-)native English speakers. As a base, the annotators used the \emph{Silver} dataset, selecting only correctly-formed sentences, thus reducing entry size to 1,551 and assuring the best quality \cite{phelps-2022-drsphelps}. Each data record was annotated by two annotators, with typical agreement being about 80\% 
. In the case where the annotators disagreed on an entry, a third made the decision.

\begin{table}[htbp]
  \begin{tabularx}{\linewidth}{>{\raggedright\arraybackslash}p{2cm}>{\centering\arraybackslash}Xl}
    \toprule
    \multirow{2}{*}{\textbf{Dataset}} & \textbf{Size} & \textbf{Quality} \\
    & \textbf{(Approx.)} & \\
    \midrule
    \textbf{Bronze}    & $\sim$1,400,000 & Low     \\
    \textbf{Silver 10} & $\sim$100,000   & Moderate   \\
    \textbf{Silver 5}  & $\sim$50,000    & Moderate   \\
    \textbf{Silver 1}  & $\sim$10,000    & Moderate   \\
    \textbf{Gold}      & $\sim$1,500     & High    \\
    \textbf{SemEval}   & $\sim$4,700     & High    \\
    \bottomrule
  \end{tabularx}
  \vspace{0.1ex}
  \caption{\centering Experimental dataset summary, \textit{SemEval} dataset included for comparison.}
  \label{tab:dataset-summary}
\end{table}

\section{Modeling}
\label{sec:Models}

\subsection{Base Model}

A similar base model to the SemEval 2022 Task 2 Subtask B baseline \cite{tayyar-madabushi-etal-2022-semeval} is used. The models are based on a pre-trained mBERT model \cite{devlin-etal-2019-bert}. MWEs in the input are represented by randomly-initialised tokens (rather than, for example, averaging the embeddings of component words) in order to avoid invalid compositional assumptions. Multiple negatives ranking loss and triplet loss are used during training in place of the baseline's cosine similarity loss, as these were shown by \citet{liu-etal-2022-ynu} to improve performance.

\subsection{Data Augmentation}

Some entrants to SemEval 2022 Task 2 \cite{tayyar-madabushi-etal-2022-semeval} improved their results by using data augmentation \citep{chu-etal-2022-hit, hauer-etal-2022-ualberta}. For enhancement of our \emph{Gold} dataset we adopt three different approaches, implemented using the NLPAug library 
\cite{ma2019nlpaug}:
        
\begin{itemize}
    \item \textbf{SpellingAug} - Substitute words using a spelling mistake dictionary;  
    \item \textbf{WordEmbsAug} - Leverage word2vec embeddings; 
    \item \textbf{TdidfAug} - Determine which word is inserted or replaced based on a TF-IDF model.
\end{itemize}

Spelling Augmentation was selected as a basic approach, introducing common spelling mistakes \cite{kulkarni_2022_natural}. The more complex approaches include using pre-trained NLP models utilising word2vec \cite{word2vec} and TF-IDF scores \cite{10.21203}. This allows insertion or replacement of likely adjacent words in sentences, rather than random augmentation.

    \subsection{Model Enhancements}
    \label{sec:ModelEnhancements}
    
The proposed model enhancements focus on incorporation of `knowledge' from one of two sources: local context (using the surrounding sentences in the input text in addition to the target sentence) and global context, adding external knowledge that is not present in the text but that a human reader could be able to infer.
\hyperref[table:ModelEnhancement]{Table \ref*{table:ModelEnhancement}} outlines the training input in each of the configurations.

\begin{table*}[htbp]
\centering
\begin{tabular}{p{0.15\linewidth} || p{0.21\linewidth} | p{0.32\linewidth} | p{0.17\linewidth} } 
  & \textbf{Local} & \textbf{Global (ParaCOMET)} & \textbf{Global (Gloss)} \\
 \hline\hline
  \textbf{Sentence-Wide Context} & Target Sentence & Target Sentence + ParaCOMET from Target Sentence & Target Sentence + Gloss from Idiom \\ 
 \hline
  \textbf{Paragraph-Wide Context} & \raggedright Target + Surrounding Sentences & Target + Surrounding Sentences + ParaCOMET from Target and Surrounding Sentences & N/A \\
\end{tabular}
\caption{\centering Overview of model enhancement and context configurations.}
\label{table:ModelEnhancement}
\end{table*}

    \subsection{Local Context}

Input to the model is expanded to include both the previous and the following sentence (where available), in addition to the target sentence. The three sentences are simply concatenated into a single text input. The addition of this extra context is likely to have mixed effects depending on the input; sentences of an entirely different topic could be concatenated together into one input, and it is unclear whether this will be more useful or harmful \citep{tayyar-madabushi-etal-2021-astitchinlanguagemodels-dataset, chakrabarty-etal-2022-rocket}.

    \subsection{External Knowledge}
       \subsubsection{ParaCOMET}
        
The COMET model, and its discourse-aware extension ParaCOMET, have been used \cite{chakrabarty-etal-2022-rocket} to return inferences focusing on specific entity relation types, which are then passed into a downstream model. A similar approach was employed in our experiments, feeding the target sentence(s) into ParaCOMET to generate external context. The ParaCOMET outputs are further concatenated with the input text.

        \subsubsection{Glosses}
        
\begin{figure}[htpb]
    \centering
    \includegraphics[width=0.7\columnwidth]{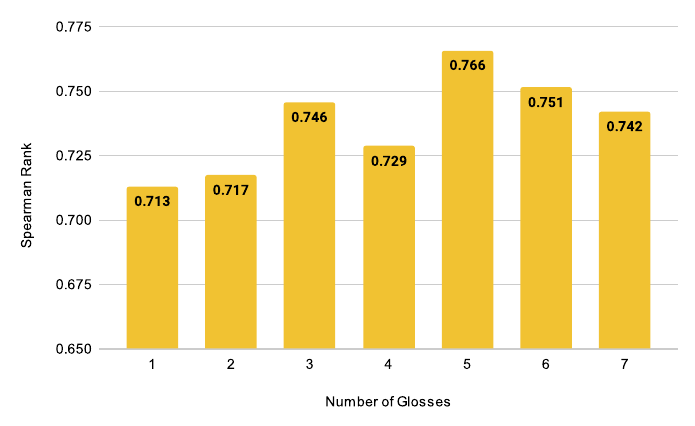}
    \caption{\centering Performance with differing number of glosses, $n$, on \emph{Silver 1} dataset.}
    \label{fig:NumberofGlosses}
\end{figure}

\citet{hauer-etal-2022-ualberta} showed that glosses can improve performance for Subtask A (idiomaticity detection), and a similar approach is taken here for Subtask B (idiom representation). For each component word of the MWE, up to $n$ glosses are retrieved from WordNet \cite{wordnet_1998} and appended to the input sentence.
The number of glosses hyperparameter, $n$, was tuned on the \emph{Silver 1} dataset, and we find 5 to be the optimal value \hyperref[fig:NumberofGlosses]{(Figure \ref*{fig:NumberofGlosses}}).

    \subsection{Evaluation}
    \label{sec:Evaluation}
    
The three model enhancement variations are all trained using our data (sub)sets, i.e. the \emph{Gold}, \emph{Silver 1}, \emph{Silver 5} and \emph{Silver 10} 
(\hyperref[tab:dataset-summary]{Table \ref*{tab:dataset-summary}})
. All of the models and their configurations are evaluated on the English subset of the SemEval evaluation dataset, using their provided scores as a baseline. Embeddings are compared using cosine similarity as in Sentence-BERT \cite{reimers-gurevych-2019-sentence} to solve the STS task, which specifically aims to evaluate whether the model actually encodes idiom meanings correctly \cite{tayyar-madabushi-etal-2022-semeval}.

\section{Results}
\label{sec:Results}

All results are presented in terms of the Spearman rank correlation \cite{DBLP:journals/corr/abs-1911-11750} between output STS scores for sentences containing idiomatic expressions and the same sentences when MWEs are replaced with non-idiomatic paraphrases; this is the same evaluation metric used to evaluate submissions to SemEval 2022 Task 2 Subtask B \cite{tayyar-madabushi-etal-2022-semeval}.

    \subsection{Dataset Factors}
        \subsubsection{Automatic Generation Evaluation}
        \label{sec:Automatic Generation Evaluation}

To evaluate how the \emph{Silver} and \emph{Gold} data set generation methods affect the quality of data, two more datasets were created by randomly sampling the \emph{Bronze} dataset down to the size of the \emph{Silver 10} and \emph{Gold} sets.

\begin{figure}[htpb]
    \centering
    \includegraphics[width=0.7\columnwidth]{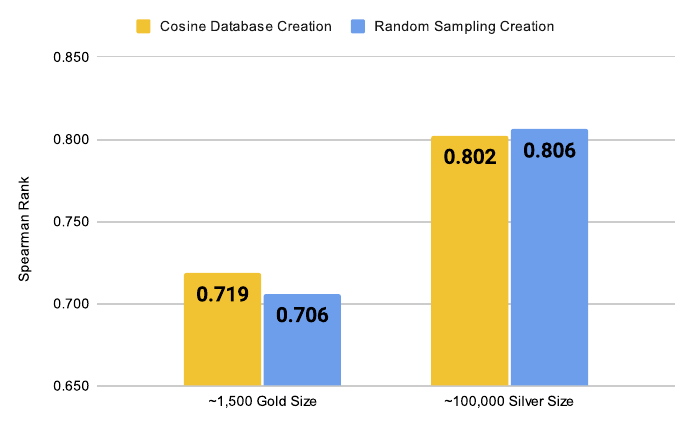}
    \caption{\centering Performance of \emph{Gold} and \emph{Silver 10} datasets against randomly-sampled \emph{Bronze} datasets of the same size.}
    \label{fig:GoldAndSilverBronzeQualityComparison}
\end{figure}

The results are presented in \hyperref[fig:GoldAndSilverBronzeQualityComparison]{Figure \ref*{fig:GoldAndSilverBronzeQualityComparison}}, where in the \emph{Gold} versus randomly sampled \emph{Bronze} case, we see a minor improvement in performance, indicating some increase in dataset quality. This is to be expected as the \emph{Gold} dataset creation method is manual and thus assures a high level of item quality.
For the \emph{Silver} comparison, there is negligible difference in performance. This suggests that the method of creating the \emph{Silver} set is insufficient to meaningfully improve the quality of the dataset. This could indicate that the quality of the \emph{Bronze} set is already high enough that it is difficult to further improve it without manual effort, weaknesses in our choice of the SemEval 2022 Task 2 training data as a high-quality reference, or flaws in the filtering method.

Despite a slight increase in quality in the \emph{Gold} dataset, it still was outperformed by the much larger, lower-quality, \emph{Silver} set. This indicates that quantity in this case seems to be the most important factor.

        \subsubsection{Data Augmentation}

\begin{figure}[htpb]
    \centering
    \includegraphics[width=0.7\columnwidth]{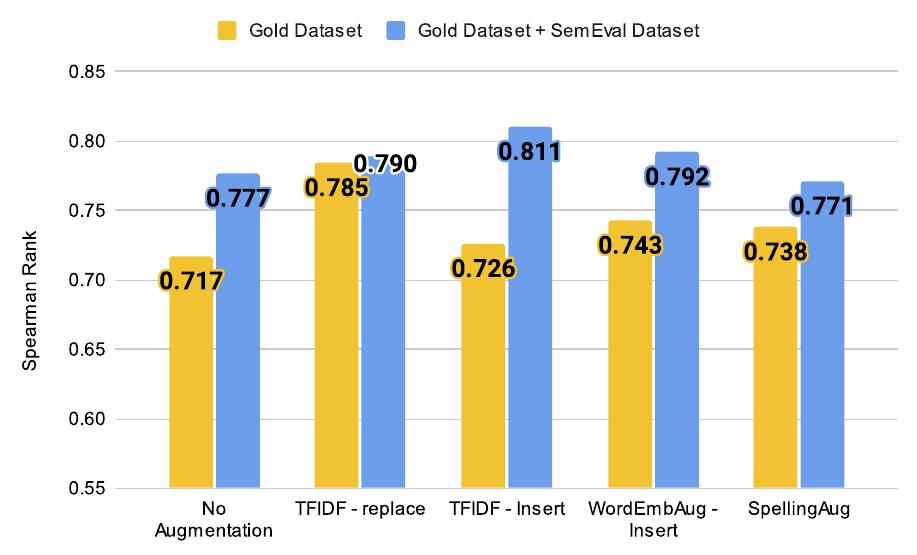}
    \caption{\centering Impact of data augmentations applied to the \emph{Gold} and \emph{Gold + SemEval} datasets.}
    \label{fig:DataAugmentation}
\end{figure}

\hyperref[fig:DataAugmentation]{Figure \ref*{fig:DataAugmentation}} presents the results models trained using the data-augmented \emph{Gold} and \emph{Gold} in combination with SemEval datasets. In each case, the augmented dataset is twice the size of the unagumented set, which is included for comparison. Each model was trained for 4 epochs and used multiple negatives ranking and triplet loss.

All but one augmentation type (spelling errors) yielded a modest improvement over the baseline. TF-IDF insertion had the biggest impact with the \emph{Gold} and \emph{SemEval} combined dataset, whereas TF-IDF replace seemed to perform better on smaller data. 
Although the dataset size is doubled via an automatic process, the performance achieved indicates that the quality remains and the larger datasets are usable; data augmentation can provide an increase in data volume while maintaining overall quality.

    \subsection{External Knowledge Factors}
        \subsubsection{Local Context}

\begin{figure}[htpb]
    \centering
    \includegraphics[width=0.7\columnwidth]{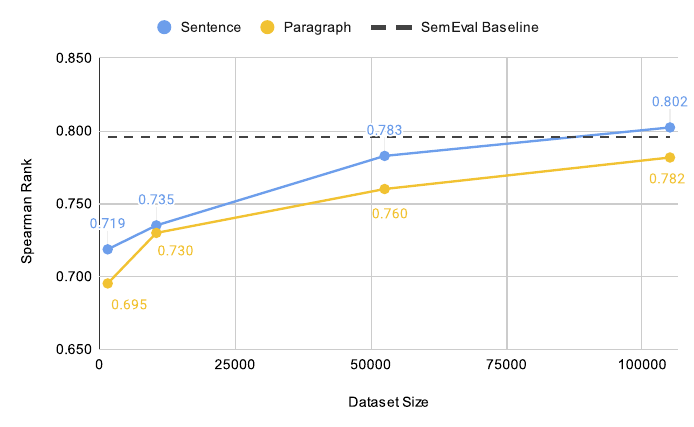}
    \caption{\centering Performance of sentence and paragraph context baseline models as dataset size increases.}
    \label{fig:BaseResultsSemEvalBaseline}
\end{figure}

\hyperref[fig:BaseResultsSemEvalBaseline]{Figure \ref*{fig:BaseResultsSemEvalBaseline}} shows the performance of our baseline models using only the target sentence and with the addition of the preceding and following sentences (where available). We observe an 
increase in performance as the dataset size increases. 

Additionally, the performance of the paragraph-wide context models is always lower than the sentence-only context. It is possible that the adjacent sentences do not contain information related to the current idiom, and as such, simply distract the model. Another possibility is that this results from the mismatch between model training and evaluation data.  As the SemEval evaluation set contains only the target potentially idiomatic sentences, models may  not be able to benefit from access to the surrounding sentences. 

        \subsubsection{Gloss Models}

\begin{figure}[htpb]
    \centering
    \includegraphics[width=0.7\columnwidth]{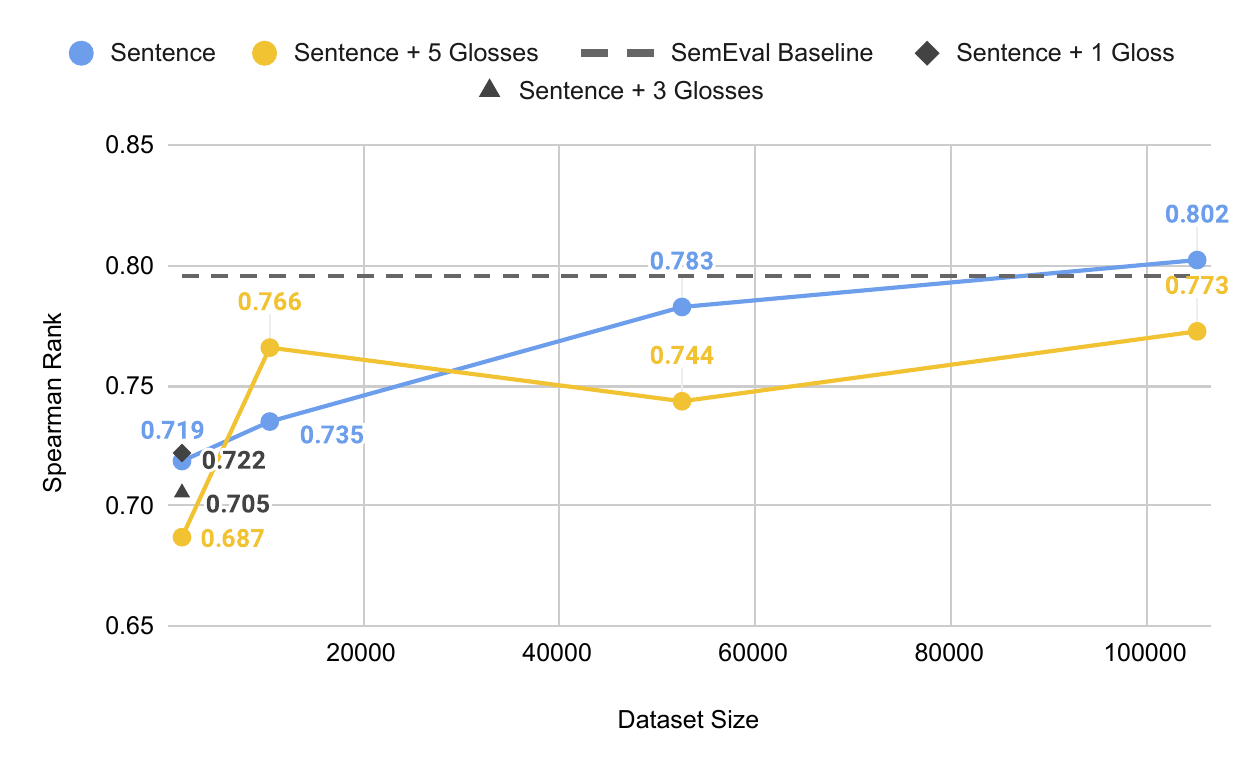}
    \caption{\centering Performance comparison of gloss and sentence-only models across datasets.}
    \label{fig:GlossPerformance}
\end{figure}

The gloss models were trained using $n=5$ glosses for 15 epochs on each dataset apart from the \emph{Silver 10} dataset, which concluded at 5 epochs. Each of these gloss models is compared to their sentence-only model ($n=0$) counterpart in \hyperref[fig:GlossPerformance]{Figure \ref*{fig:GlossPerformance}}.

The gloss model shows improvement over the sentence-only model only on the \emph{Silver 1} dataset. This could be caused by the choice of the number of glosses hyperparameter, which was tuned on the \emph{Silver 1} data. In order to test this theory, models were trained on the \emph{Gold} dataset with $n=1,3$. The $n=1$ model showed marginal improvement over the sentence model, though this is unconvincing evidence that hyperparameter selection is the sole reason for the performance on the \emph{Silver 1} dataset. 

\begin{figure}[htpb]
    \centering
    \includegraphics[width=0.7\columnwidth]{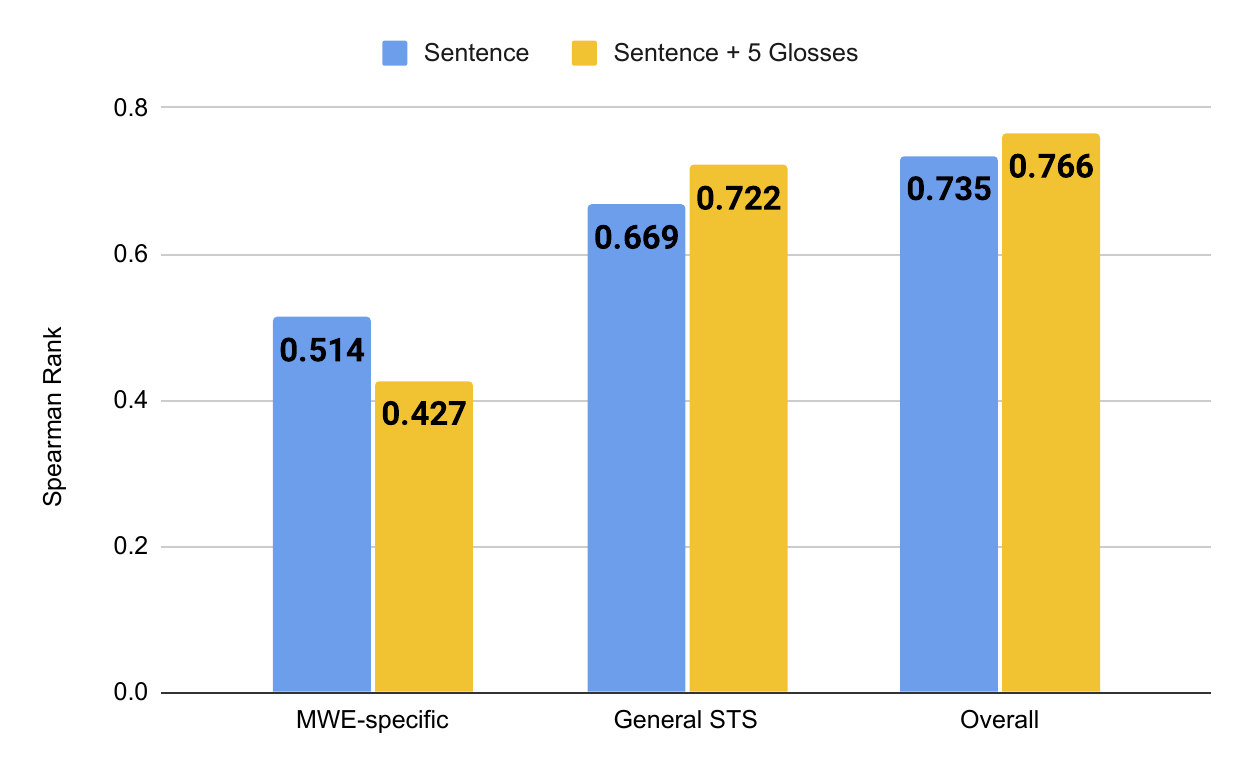}
    \caption{\centering Model performance (trained on \emph{Silver 1}) on general STS  and MWE-specific evaluation.}
    \label{fig:GlossPerformanceIndividual}
\end{figure}

A comparison is made between the sentence and gloss models' performance on the general STS and MWE-specific portions of the evaluation dataset \hyperref[fig:GlossPerformanceIndividual]{(Figure \ref*{fig:GlossPerformanceIndividual})}. The gloss model performs better on the general STS data, whereas the sentence model performs better on the MWE-specific data. This suggests that using glosses for MWE component words is beneficial for the sentence similarity task but does not improve the models' representation of idioms.

        \subsubsection{ParaCOMET Model}
        
We investigated using sentence-only and paragraph-wide local context and ParaCOMET knowledge, generating either 5 or 12 inferences. Inferences generated were of the types \emph{xNeed}, \emph{xIntent}, \emph{xWant}, \emph{xReact}, \emph{xEffect} and \emph{xAttr} as suggested by \citet{chakrabarty-etal-2022-rocket}.

\begin{figure}[htpb]
    \centering
    \includegraphics[width=0.7\columnwidth]{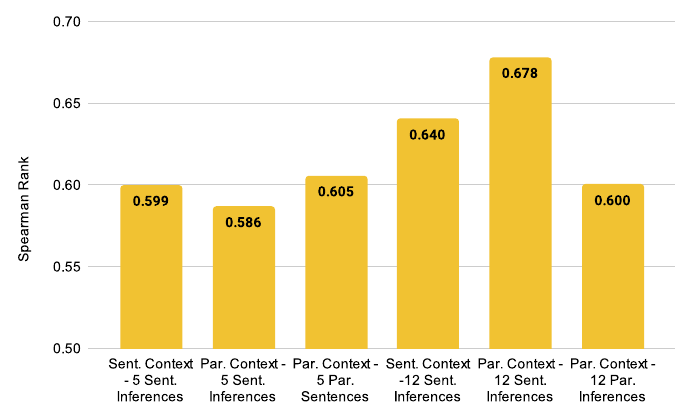}
    \caption{\centering Performance of each ParaCOMET knowledge enhancement combination, training on the \emph{Silver 1} dataset.}
    \label{fig:ParaCOMETOnSilver1}
\end{figure}

Focusing on each model's performance on the \emph{Silver 1} dataset, a medium-sized dataset that is not handpicked, we obtained the highest performance when appending the surrounding sentences to the target sentence along with 12 inferences generated from the target sentence, as outlined in \hyperref[fig:ParaCOMETOnSilver1]{Figure \ref*{fig:ParaCOMETOnSilver1}}.

\begin{figure}[htpb]
    \centering
    \includegraphics[width=0.7\columnwidth]{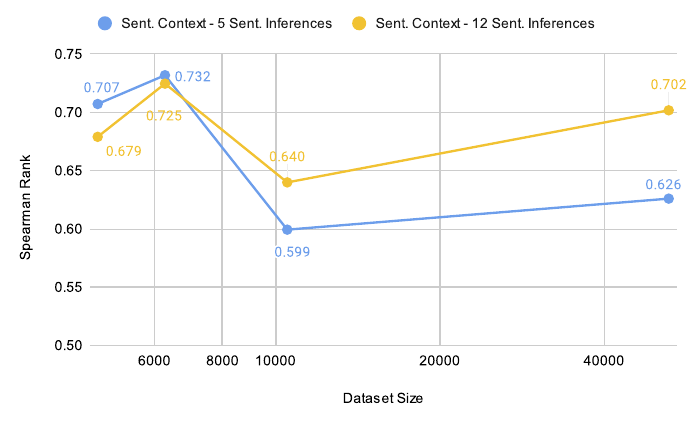}
    \caption{\centering Performance of two ParaCOMET-enhanced models across datasets.}
    \label{fig:ParaCOMETNumInferences}
\end{figure}

Results across the four datasets for models using sentence-only local context with either 5 or 12 sentence inferences are presented in \hyperref[fig:ParaCOMETNumInferences]{Figure \ref*{fig:ParaCOMETNumInferences}}. 

We observe a significant drop in performance as we move from manually-collected \emph{Gold} to an automatically scraped \emph{Silver} dataset. The results also suggest that a lower number of inferences is more beneficial when combined with a smaller, high-quality dataset, while a higher number of inferences helps when the dataset quality is lower.

\section{Discussion}
\label{sec:Discussion}

\begin{figure}[htpb]
    \centering
    \includegraphics[width=0.7\columnwidth]{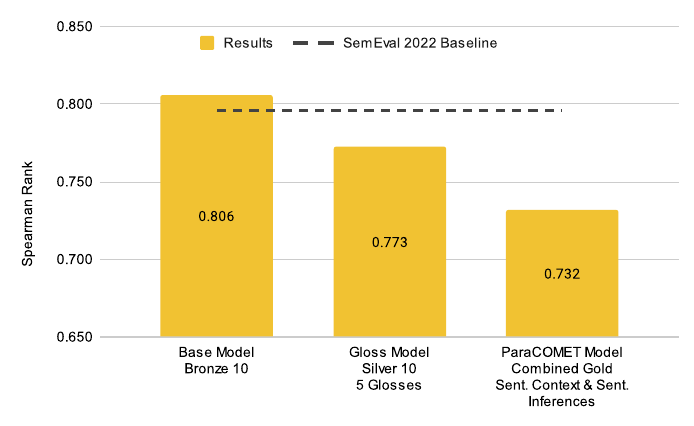}
    \caption{\centering Best-performing model configurations.}
    \label{fig:OverallResults}
    \end{figure}

    \hyperref[fig:OverallResults]{Figure \ref*{fig:OverallResults}} highlights the best-performing configurations of the three constructed models. The base model without any modifications is the only one to outperform the SemEval baseline. However, note that these models are trained without direct access to the SemEval training set -- our results when combining our data with the SemEval training (see \hyperref[fig:DataAugmentation]{Figure \ref*{fig:DataAugmentation}}) suggest that this would yield further improvements.
    
    Our hypothesised rise in performance as dataset quality increases is generally not observed. However, we expect that the results are impacted by the small difference in quality between the datasets 
    , especially \emph{Bronze} and \emph{Silver}. Furthermore, the \emph{Gold} dataset, which has a more significant quality increase, might not have been of sufficient size, as its counterpart from the SemEval Task 2 is three times larger. 
    It is worth noting, however, that in the case of context-enhanced models, performance did increase with dataset quality.
    
   It appears that providing the model with surrounding sentence context is \textbf{not} generally useful for all model configurations, with or without enhancement. It can also be observed that across context-enhanced models, context specific to the idiomatic MWE plays the most significant part, as the best-performing models occur when they are constrained to the target sentence (\hyperref[fig:ParaCOMETOnSilver1]{Figure \ref*{fig:ParaCOMETOnSilver1}}) or do not have too much possibly distracting additional information attached (\hyperref[fig:NumberofGlosses]{Figure \ref*{fig:NumberofGlosses}}, \hyperref[fig:GlossPerformance]{Figure \ref*{fig:GlossPerformance}}). This leads to the intuition that if context is to be valuable for idiom representation, it should be as specific and relevant as possible to the MWE in question.

\section{Conclusions}
\label{sec:Conclusions}

    In this work we introduce the Noun Compound Synonym Substitution in Books (\textbf{NCSSB}) datasets, and use them to investigate the impact of factors like data quality and quantity in modeling idiomatic expressions, as well as examining the inclusion of different types of contextual information.
    
    For identification of idiomatic language, a balance between dataset quality and quantity needs to be reached: a small dataset of high quality might not be enough to fully enable the model to accurately represent idiomatic MWEs. Nonetheless, quality seems to have a stronger impact than quantity when using knowledge-enhanced models.
    
    
    While the inclusion of local context from adjacent sentences led to decreased model performance in all scenarios, the use of external knowledge from glosses and ParaCOMET inferences led to better performance. These results suggest that for models where targeted external knowledge is available, increases in performance can be obtained with small but high-quality datasets.  
    For models with no knowledge enhancement, larger training datasets led to better performance. Therefore, dataset quality and quantity seem to interact with characteristics of the model and access to contextual information. A balance including high quality data and access to context is at the core of idiomatic language understanding in natural language models.

\section*{Acknowledgments}
This work was in part supported by the Centre for Doctoral Training in Speech and Language Technologies (SLT) and their Applications funded by UK Research and Innovation [grant number EP/S023062/1]. The work was also partly supported by The Healthy Lifespan Institute (HELSI) at The University of Sheffield, UK EPSRC grants EP/T517835/1 and EP/T02450X/1, by the Turing Institute Fellowship and by the UniDive COST Action.

\bibliographystyle{unsrtnat}
\bibliography{custom,lrs,acl_anthology}

\end{document}